\documentclass[lettersize,journal]{IEEEtran}
\usepackage{amsmath,amsfonts}
\usepackage{algorithmic}
\usepackage{algorithm}
\usepackage{array}
\usepackage{ulem}
\usepackage{soul}
\usepackage{xcolor}
\usepackage[caption=false,font=normalsize,labelfont=sf,textfont=sf]{subfig}
\usepackage{textcomp}
\usepackage{stfloats}
\usepackage{url}
\usepackage{verbatim}
\usepackage{graphicx}
\usepackage{cite}
\usepackage{amsmath,amsfonts}

\usepackage{booktabs}
\usepackage{mathrsfs}
\usepackage{diagbox}
\usepackage{threeparttable}
\usepackage{multirow}
\usepackage[T1]{fontenc}
\usepackage{amssymb}
\usepackage{makecell}

\sethlcolor{yellow}

\soulregister{\cite}{7}
\soulregister{\ref}{7}
\soulregister{\pageref}{7}

\soulregister{\cite}{7}
\soulregister{\ref}{7}
\soulregister{\pageref}{7}

\begin{document}

\title{P2LHAP:Wearable sensor-based human activity recognition, segmentation and forecast through Patch-to-Label Seq2Seq Transformer}

\author{Shuangjian Li, Tao Zhu,~\IEEEmembership{Member,~IEEE},
	Mingxing Nie, Huansheng Ning,~\IEEEmembership{Senior Member,~IEEE,} Zhenyu Liu and  Liming Chen  ~\IEEEmembership{Senior Member,~IEEE}
 

\thanks{This work was supported in part by the National Natural Science Foundation
of China (62402209, 62006110), the Natural Science Foundation of Hunan Province (2024JJ7428, 2023JJ30518). (Corresponding author: Tao Zhu.)}
\thanks{Shuangjian Li, Tao Zhu, Mingxing Nie and Zhenyu Liu are with the School of Computer Science, University of South China, 421001 China (e-mail: sjli@stu.usc.edu.cn, tzhu@usc.edu.cn, niemx@usc.edu.cn, lzy@usc.edu.cn). Liming Chen was with School of Computer Science and Technology, Dalian University of Technology, China. Huansheng Ning was Department of Computer \& Communication Engineering, University of Science and Technology Beijing, 100083 China (email: l.chen@ulster.ac.uk, ninghuansheng@ustb.edu.cn).}
}
\markboth{Journal of \LaTeX\ Class Files,~Vol.~14, No.~8, August~2021}%
{Shell \MakeLowercase{\textit{et al.}}: A Sample Article Using IEEEtran.cls for IEEE Journals}

\maketitle

\begin{abstract}
Traditional deep learning methods struggle to simultaneously segment, recognize, and forecast human activities from sensor data. This limits their usefulness in many fields such as healthcare and assisted living, where real-time understanding of ongoing and upcoming activities is crucial. This paper introduces P2LHAP, a novel Patch-to-Label Seq2Seq framework that tackles all three tasks in a efficient single-task model. P2LHAP divides sensor data streams into a sequence of "patches", served as input tokens, and outputs a sequence of patch-level activity labels including the predicted future activities. A unique smoothing technique based on surrounding patch labels, is proposed to identify activity boundaries accurately. Additionally, P2LHAP learns patch-level representation by sensor signal channel-independent Transformer encoders and decoders. All channels share embedding and Transformer weights across all sequences. Evaluated on three public datasets, P2LHAP significantly outperforms the state-of-the-art in all three tasks, demonstrating its effectiveness and potential for real-world applications.
\end{abstract}


\section{Introduction}
\IEEEPARstart{T}{he} intersection of deep learning and wearable sensor technology for real-time human activity recognition has emerged as a prominent research focus in recent years. This area holds significant promise for applications in medical health, assisted living, smart homes, and other domains. In various practical scenarios, there is a need for perceptual techniques to delineate activity boundaries, identify activity categories, and predict forthcoming activities from the continuous stream of sensor data. As a result, the field of human activity perception encompasses three pivotal tasks: human activity segmentation (HAS)\cite{10210541}, activity recognition (HAR)\cite{chen2023hmgan}, and activity forecast (HAF)\cite{park2022enhanced}. While extensive research\cite{saleem2023toward, 8727452} has addressed each of these tasks individually, limited attention has been devoted to simultaneously addressing all three tasks in a cohesive framework.


Several approaches exist for concurrently achieving activity segmentation, recognition, and forecast, as illustrated in Fig. 1. The first approach, shown in Fig. 1(a), employs three independent single-task learning models in parallel. These models output the category of the current activity, the boundary of the current activity, and the category of the next activity. However, this method has drawbacks, including disparate data annotation requirements for different models, the need for additional policies to ensure compatibility among model outputs, and a lack of information sharing between models, leading to inefficient use of computing resources.

A second approach involves multi-task learning in Fig. 1(b), recent advancements in multi-task Human Activity Recognition learning frameworks \cite{duan2023multi, xia2022boundary, kim2023multi} leverage the interdependencies among tasks to facilitate data sharing and encoder learning, enabling the extraction of common data representations. Following this, multiple lightweight single-task models are employed. While this approach often enhances multitasking effectiveness, it introduces increased model complexity compared to single-task models. Managing this complexity may require additional computational resources and careful architectural design. Simultaneously, it is crucial to ensure consistency in the model output representation across different tasks. Our objective is to standardize the output representation for improved task-specific adaptability. Furthermore, these multitask models can be more sensitive to hyperparameter choices, requiring meticulous tuning to achieve optimal performance across all tasks.


Different from the first two methods, as shown in Fig. 1(c), this paper proposes a novel Patch-to-Label Seq2Seq framework that tackles all the three human activity perception tasks in a single-task model, named P2LHAP. P2LHAP adopts the encoder-decoder framework, the input is a patch sequence of sensor data, and the output is a sequence of patch-level activity label, where a patch is a short segment of sensor data points. With patch-level labels, it can simply Merge adjacent identical labels to determine the category of the active boundary. In addition, using the encoder-decoder architecture of seq2seq, it can easily output future activity token sequences, making it simple to predict activity label. The architecture handling capability of variable-length sensor activity sequences is excellent, and it exhibits high adaptability to sequences of indefinite length, comprehensively considering all information within the input sensor signal sequence and effectively retaining the contextual information of the sensor data stream.

Using patches offers several advantages over individual data points and sliding windows. Compared to using single data points\cite{yao2018efficient, xia2022boundary}, patch-based methods help mitigate the effects of noise by enabling feature extraction within localized intervals\cite{Yuqietal-2023-PatchTST}. Most existing activity recognition research employs one or more fixed-length sliding Windows to segment the data stream, under the assumption that each window captures a complete activity\cite{duan2023multi}. However, because activities have varying durations, it is very difficult to set an appropriate window length. If it is too short, the window may not capture sufficient information; if it is too large, may encompass multiple activities of different categories, both of which will reduce the recognition accuracy \cite{qian2021weakly}. The patching strategy can circumvent this problem well, because we assume that an activity can contain any number of patches, and encoder-decoder architectures such as Tranfomer are good at handling variable-length sequences, making it feasible to identify activities of arbitrary duration.
\begin{figure*}[htb]
	\centering
	\includegraphics[width=\textwidth]{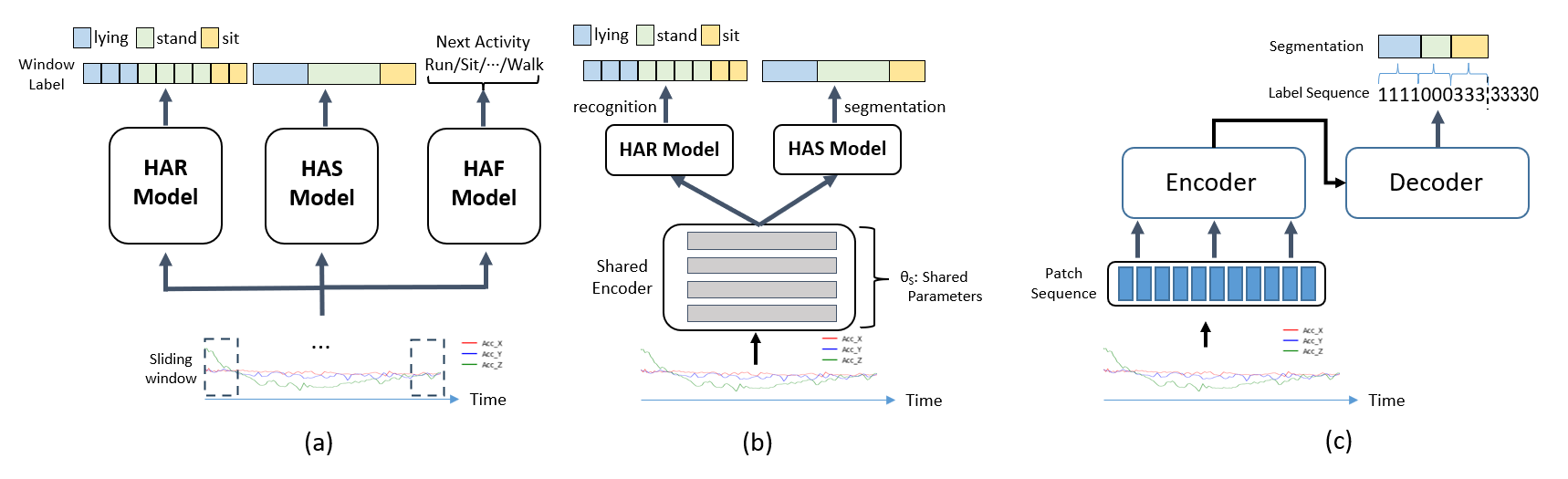}
	\label{segmentmodel}

        \caption{There are different architectures for multitasking approaches as follows: (a) The data is partitioned using sliding windows, followed by separate training of the respective models. These trained models are subsequently utilized for recognition, segmentation, and forecast. (b) Multi-task methods learn models for multiple tasks through a shared encoder and generate separate results for each task. (c) We propose a multi-task P2L Seq2Seq model specifically designed to generate patch-level active label sequences.}
\end{figure*}

Although Transformers have achieved great results in the fields of NLP\cite{vaswani2017attention} and CV\cite{dosovitskiy2020vit}, recent research shows that classic Transformers do not work well on multi-variable long sequence time series prediction problems\cite{zeng2023transformers}. This may be because multi-variable features will interfere Extraction of long-term sequence features\cite{Yuqietal-2023-PatchTST}. Limited training data is also a issue, in contrast to image and text data, the scarcity of activity recognition datasets can impact the Transformer model's performance. Despite their modeling flexibility surpassing that of convolutional neural networks (CNNs) or recurrent neural networks (RNNs), Transformers can pinpoint relevant information at any position. Consequently, substantial datasets are typically necessary to realize their full potential. Human activity perception based on wearable devices is essentially a long-sequence multi-variable time series analysis problem. Sensor data includes signals from accelerometers, gyroscopes, magnetometers, etc. Inspired by\cite{Yuqietal-2023-PatchTST}, instead of using the classic Transformer architecture, we introduce a channel-independent Transformer architecture as the encoder-decoder of P2LHAP, each patch only contains data from a single wearable sensor channel information, each sensor signal sequences within the dataset undergoes processing independently within the Transformer model, allowing for the learning of diverse attention patterns tailored to each sequence. In contrast to the conventional architecture where all sequences are subject to the same attention mechanism, this individualized approach prevents potential interference caused by disparate behaviors within multi-channel sensor data. Moreover, by adopting a channel-independent strategy, the adverse impact of noise prevalent in certain sensor signal sequences can be mitigated as it no longer gets projected onto other sequences within the embedding space.

we utilize the original sensor data sequence along with the patch features obtained from the encoder to compute patch-level representations through an attention network, which are then used to predict activity labels for each patch. However, due to the variable size of the patch, the ultimate active label sequence may suffer from over-segmentation. To address this issue, we devised a smoothing algorithm for the final Patch-level activity label sequence, aimed at mitigating the over-segmentation phenomenon. This algorithm considers the number of activity categories surrounding the patch to adjust the current patch-level activity label. It then establishes activity boundaries based on the label range, significantly reducing boundary ambiguity in the activity signal.

The main contributions of this paper can be summarized as follows:
\begin{itemize} 
    \item  We proposes a novel Patch-to-Label Seq2Seq framework that tackles all the three human activity perception tasks in a single-task model, named P2LHAP. To the best of our knowledge, our work constitutes the pioneering effort in unifying activity segmentation, recognition, and forecast within a single framework.
    \item  We employ patches to address the challenge of insufficient information from large sliding windows or individual data points, and our method, utilizing a seq2seq encoder-decoder architecture, readily generates future activity token sequences, making it straightforward to forecast activity labels.
    \item Each patch of our approach  only contains data from a single wearable sensor channel information, each sensor signal sequences within the dataset undergoes processing independently within the Transformer model, allowing for the learning of diverse attention patterns tailored to each sequence. 
    \item Designed a smoothing method to analyze the number distribution of category labels surrounding each active label. This information is used to update the current active label, effectively reducing the occurrence of over-segmentation errors and providing smoother predictions within the same activity category.
        \item The proposed P2LHAP method has been evaluated through experiments conducted on three widely used benchmark datasets for activity recognition, segmentation and forecast. The experimental results demonstrate that our approach surpasses existing methods in terms of performance.
\end{itemize}

\section{Related Works}
This section provides a comprehensive review of the progress made in related research on multi-task activity recognition, segmentation, and forecast utilizing sensor signals.
\subsection{Sensor-based activity recognition and segmentation tasks}

Sensor-based human activity recognition primarily relies on IMU sensors in wearable devices like smartwatches, smart insoles, and exercise bands. Traditional approaches, such as fixed-size sliding windows and dense labeling, have demonstrated strong performance \cite{chen2021deep}. Recent studies have employed diverse techniques. \cite{yi2023human} distinguished static and dynamic activities using statistical techniques, then classified specific activities with random forests (RF) and CNNs. In\cite{10179199}, NAS was used to investigate effective deep learning (DL) architectures for smartphone inference. \cite{10348511} introduced a lightweight CNN-BiGRU model, leveraging inertial sensor data from smartwatches and smartphones, achieving satisfactory results. However, the inherent uncertainty in activity duration, these methods will lead to multi-class window problems, where multiple activities may coexist within a window, introducing noise and negatively impacting model recognition performance. \cite{9055403} employed an LSTM-based method to identify fine-grained patterns using high-level features from sequential motion data, \cite{9732352} introduced a deformable convolutional network for activity recognition from complex sensory data, and \cite{mim2023gru} presented a GRU-INC model that initializes attention-based GRUs to effectively leverage spatiotemporal information in time series. Additionally, \cite{challa2022multibranch} proposed a multi-branch CNN-BiLSTM network that extracts features from raw sensor data with minimal preprocessing. \cite{zhang2021conditional} further developed Conditional-UNet, which models conditional dependencies between dense labels for coherent HAR. Current advancements in activity segmentation and recognition, such as \cite{10210541, xia2022boundary}, leverage dense labeling methods. However, these methods rely solely on individual sensor samples as neural network inputs, and will obtain different representations depending on the task. Our proposed approach employing a patch block, situated between the scales of a large sliding window and a single sample point, for data segmentation. We then utilize a Seq2Seq architecture to unify activity segmentation and recognition tasks, obtaining a unified activity label sequence representation to perform different tasks while mitigating the challenges associated with multi-class windows and over segmentation.

\subsection{forecast of activities using wearable sensors}

Sensor-based human activity recognition systems have gained popularity with the widespread use of wearable devices in daily life. However, there remains a significant gap in research focusing on sensor-based activity forecasting. Recent studies have introduced attention models to predict multivariate motion signals from IMU sensors, thereby preserving pattern and feature information across multiple channels. Predictive models for these time series data can be utilized for human activity forecast by generating and classifying future activity signals. In \cite{s21248270}, a combination of a neural network and Fourier transform was employed to convert the sensor activity data into an image format for predicting future signals. In \cite{biology9120441} introduced the SynSigGAN model, which utilizes bidirectional grid long short-term memory as the generator network of the GAN model, and a convolutional neural network as the discriminator network to generate future signals. Furthermore, recent work \cite{10.1007/978-3-031-09342-5_13} proposes a Transformer-based GAN that can successfully generate real synthetic time series data sequences of arbitrary length, similar to real data sequences. The GAN model’s generator and discriminator are built using a pure Transformer encoder architecture, which outperforms traditional GAN structures based on RNN or CNN models. Despite these advancements, existing methods predominantly rely on training models to forecast and emulate future sensor activity signals, subsequently employing classification algorithms to derive activity categories from the forecasted signal sequences. Any inaccuracies in the forecasted sensor signals inevitably impact the final classification outcome. In contrast, our proposed P2LHAP architecture bypasses this intermediate step by directly forecasting future activity category token sequences, thereby enhancing forecast accuracy. An additional study \cite{Yuqietal-2023-PatchTST} presented the PatchTST time series forecast model, which is based on channel independence and patching, and demonstrated promising results. While there has been progress in forecasting multivariate sensor time series data, research on multi-category forecast of human activities based on multi-sensor signal channels remains limited.

\subsection{Sensor-based multi-task learning framework}
Sensor-based multi-task activity recognition presents a significant challenge in the field. \cite{MARTINDALE2021250} employs recurrent neural networks to segment and identify activities and cycles using inertial sensor data. \cite{kim2023multi} proposes an efficient multi-task learning framework that utilizes commercial smart insoles. This framework addresses three health management-related problems by converting smart insole sensor data into recursive graphs and employing an improved MobileNetV2 as the backbone network. Experimental results demonstrate that this multi-task learning framework outperforms single-task models in tasks such as activity classification, speed estimation, and weight estimation. \cite{9098923} introduces a multi-task model based on LSTM for predicting each activity and estimating activity intensity using wearable sensor signals. \cite{xia2022boundary} combines the segmentation and classification tasks by passing the sensor signal through the MS-TCN module to predict sample-level labels and activity boundaries. This joint segmentation and classification approach calculates sample-level loss and boundary consistency loss to achieve improved performance. \cite{duan2023multi} proposes a novel multi-task framework called MTHARS, which leverages the dynamic characteristics of activity length to segment activities of varying durations through multi-scale window splicing. This approach effectively combines activity segmentation and recognition, leading to improved performance in both tasks. However, prior studies employed distinct backbone networks for various tasks or failed to achieve unified task representations, our proposed architecture integrates these tasks within a single model for simultaneous learning and representation. By doing so, we obtain a unified label sequence representation.

\section{Method}
\begin{figure*}[htb]
	\centering
	\includegraphics[width=\textwidth]{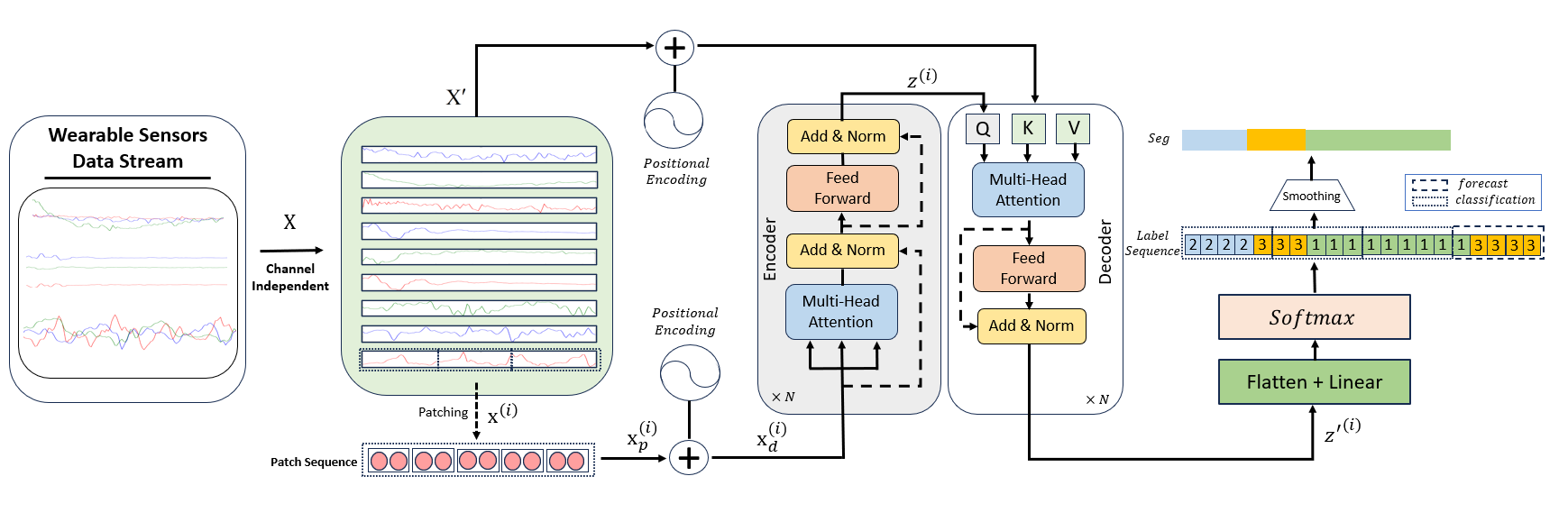}
	\caption{Overview of the Proposed P2LHAP framework}
	\label{segmentmodel}
\end{figure*}
In this section, we introduce our proposed method, which is a multi-task model for human activity recognition utilizing the Patch-to-Label Seq2Seq approach. Fig. \ref{segmentmodel} provides an overview of the model's architecture. This model enables activity recognition, segmentation, and forecast tasks on continuous sensor data streams collected from wearable devices. Our approach consists of three primary components: 1) Patch-to-Label Seq2Seq framework, 2) Patching and channel independent Transformer architecture, and 3) Activity classification, segmentation, and forecasting module.

\subsection{Patch-to-Label Seq2Seq framework}
The Patch-to-Label Seq2Seq framework we propose is shown in Fig. 2. We define the $M$-dimensional input wearable sensor sequence as $\mathbf{X}_{1:L}=[\mathbf{x}_1,\cdots ,\mathbf{x}_L],$$ $ $\mathbf{x}_t\in\mathbb{R}^{1 \times M}$, where $\mathbf{x}_t$ represents the sensor signals collected at timestamp $\textit{t}$. The activity label for each sample is denoted as $y_{1:L}=[y_1, \cdots , y_L],$ $y_t\in R^{\textit{C}}$, where $\textit{C}$ represents the number of activity categories, $\textit{t}$ is the time step, \textit{L} is the length of the wearable sensor sequence, and the length of the patch block is defined as \textit{P}. In this study, we do not employ a fixed-length sliding window to segment the sensor data. The number of patches is denoted as \textit{L}/\textit{P}. Our approach utilizes a Transformer encoder and decoder as the underlying framework.  To simultaneously process the sensor patch of each channel in parallel within the Transformer backbone, it is crucial to duplicate the Transformer weights multiple times and employ the patching operator to create a sensor sequence sample of a size $B \times M \times L$. The patch-level sequence extends to $M \times P \times N$. Reshaping the dimension sequence into a \textit{3D} tensor of size $(B \cdot M) \times P \times N$, the batch of samples serves as the input for the encoder. The purpose of the encoder is to map multiple sequences of single-channel sensor data patches into a latent representation. Following this, the decoder is used to derive the representation of the activity sequence, denoted as  $\mathbf{Z}^{' (i)} $. The ultimate output signifies the active label result obtained through the classification head, the active label sequence and the forecast of future activity label sequence of length $T_{pre}$, denoted as $\hat{y}_{pre} =\left \{  \hat{y}_1,\cdots , \hat{y}_{\textit{L}+T_{pre}}  \right \} $.

\subsection{Patching and channel independent Transformer architecture}
\subsubsection{\bf{Patching and channel independent method}}
We process each sensor signal channel independently, specifically, we segment the input sensor signal sequence $ \mathbf{X}_{1:L} $ into $M$ channel sensor signal sequences $\mathbf{x}^{(i)}$ based on its dimension $M$, $\mathbf{x}^{(i)}$ represents the sensor sequence of the $i$-th channel, $i \in \{1,2, \cdots, M\}$. We normalize each $\mathbf{x}^{(i)}$ before patching and the mean and deviation. Next, we divide $\mathbf{x}^{(i)}$ into overlapping or non-overlapping Patches and the length of the Patch block is $P$. The overlapping part is its step size, represented by $S$. By segmentation, a patch-level sensor sequence $\mathbf{x}_{p}^{(i)} \in \mathbb{R}^{1\times N}$ is obtained, where $N=\left \lfloor \frac{(L-P)}{S} \right \rfloor +2$. $\mathbf{x}_{p}^{(N)}$ is the last patch of the segmented patch sequence, this patch is learned through the previous $N-1$ patches, a sequence representation into the future is obtained, with its initial value filled by repeatedly replicating the last number of the original sequence. Finally, we add the corresponding position coding to $\mathbf{x}_p^{(i)}$ to obtain $\mathbf{x}_d^{(i)}$, and then transmit the time series of each channel to the Transformer backbone respectively.

\subsubsection{\bf{Encoder}}

We use the native Transformer encoder \cite{dosovitskiy2020vit} to model the features of patch-level sensor sequences, similar to the method used in \cite{Yuqietal-2023-PatchTST}. The Encoder module utilizes trainable linear projections to map continuous Patch blocks to a \textit{D-dimensional} Transformer latent space $\mathbf{W}_p\in \mathbb{R}^{D\times P}$. It incorporates a learnable additive positional encoding $\mathbf{W}_{pos}\in \mathbb{R}^{D\times N}$ to capture the temporal order of the patches, represented by $\mathbf{x}_d^{(i)}=\mathbf{W}_p\mathbf{x}_{p}^{(i)}+\mathbf{W}_{pos}$, where $\mathbf{x}_{d}^{(i)}\in\mathbb{R}^{D\times N}$ represents the input fed into the Transformer encoder. For each head $h=1,\cdots,H$ in the multi-head attention, the Patch blocks are transformed into query matrices $\mathbf{Q}_{h}^{(i)}=(\mathbf{x}_{d}^{(i)})^T\mathbf{W}_{h}^Q $, key matrices $\mathbf{K}_{h}^{(i)}=(\mathbf{x}_{d}^{(i)})^T\mathbf{W}_{h}^K $, and value matrices $\mathbf{V}_{h}^{(i)}=(\mathbf{x}_{d}^{(i)})^T\mathbf{W}_{h}^V $, where $\mathbf{W}_{h}^Q,\mathbf{W}_{h}^{K}\in\mathbb{R}^{D\times d_k},\mathbf{W}_{h}^{V}\in\mathbb{R}^{D\times D}$. Subsequently, the attention outputs are normalized to obtain the final attention output $\mathbf{Q}_{h}^{(i)}\in\mathbb{R}^{D\times N} :$
\begin{equation} 
    (\mathbf{Q}_{h}^{(i)})^{T} = Attention(\mathbf{Q}_{h}^{(i)},\mathbf{K}_{h}^{(i)},\mathbf{V}_{h}^{(i)})
\end{equation}
\begin{equation} 
    Attention(\mathbf{Q},\mathbf{K},\mathbf{V}) = softmax(\frac{\mathbf{Q}\mathbf{K}^T}{\sqrt{d_k} })\mathbf{V}
\end{equation}

The multi-head attention module also includes a BatchNorm layer and a feed-forward network with residual connections.
   
\subsubsection{\bf{Decoder}}
The Decoder module combines the representation $\mathbf{Z}^{(i)}=\left \{ \mathbf{z}_1,\cdots ,\mathbf{z}_p \right \} $, derived from the encoder output, with the corresponding Patch-level wearable sensor data $\mathbf{X}^{'}=\left \{ \mathbf{x}_1^{'} ,\cdots ,\mathbf{x}_p^{'}\right \} $ to generate the activity label sequence using a key-query-value attention mechanism. By multiplying the weight matrices $\mathbf{W}_k$ and $\mathbf{W}_v$ with the Patch-level wearable sensor data, we obtain representations of the key and value, denoted as $\mathbf{K}=\mathbf{W}_k\mathbf{X}^{'}$ and $\mathbf{V}=\mathbf{W}_v\mathbf{X}^{'}$, respectively. At the same time, the query vector $\mathbf{Q}$ is obtained by multiplying the feature representation $\mathbf{z}^{(i)}$ from the encoder with the weight matrix $\mathbf{W}_q$, expressed as $\mathbf{Q}=\mathbf{W}_{q}\mathbf{Z}^{(i)}$. Subsequently, the inner product of $\mathbf{Q}$ and $\mathbf{K}$ is passed through a temperature softmax layer to obtain attention scores, which are then used to weigh the summation of the value vector $\mathbf{V}$, resulting in the representation of $\mathbf{z}^{'(i)}:$
\begin{equation}
    \mathbf{z}^{'(i)}=softmax(\mathbf{Q}^T\mathbf{K}/\sigma)\mathbf{V}=f_{\theta}(\mathbf{X}^{'},\mathbf{Z}^{(i)})    
\end{equation}

Here $\sigma$ is the temperature parameter, and $f_{\theta}(\cdot)$ represents the attention mechanism network.
\subsection{Activity classification}
we employ a Flatten layer with a linear head and a Softmax layer to generate the final human activity label sequence $\hat{y}$. The representation of $\hat{y}$ is summarized as follows:
\begin{equation}
    softmax(x)=\frac{e^{x_i} }{ {\textstyle \sum_{j=1}^{n}e^{x_j}} } 
\end{equation}
\begin{equation}
    \hat{y}=argmax_i(softmax(Linear(\mathbf{z}^{'})) 
\end{equation}

During the training phase, we utilize patch-level label predictions and true labels to construct a patch-level classification loss, aiming to effectively optimize the model parameters. This loss is defined as follows:

\begin{equation}
    L_{cls}=l_p(\hat{y},y )
\end{equation}

Among them, $\hat{y}$ and $y$ represent the patch-level predicted label and the real label, respectively. Given the imbalance of human activity classes under unconstrained conditions, we employ the Softmax CrossEntropy loss function $l_p(\cdot)$ and take into account the number of effective samples for classification. The calculation formula is as follows:

\begin{equation}
    l_p(\hat{y}, y)=-\frac{1}{L} \sum_{l=1}^{L}\sum_{c=1}^{C}ylog(\hat{y} )  
\end{equation}

Where  $L$ signifies the length of input sequence, and C represents the activity class number of dataset.  In sensor-based HAR tasks, we leverage this loss function and take into account the number of effective samples to amplify the weight assigned to minority classes. This approach aims to yield fairer and more robust results for each individual class.
\subsection{Activity segmentation and smoothing strategy}
Given that the classification loss treats each patch independently, it may result in unwarranted over-segmentation errors. To achieve smoother transitions between patches and minimize over-segmentation errors, we employ the truncated mean square error as our chosen smoothing loss function. The formulation of the smoothing loss is as follows:

\begin{equation}
    L_{seg}=L_{T-MSE}=\frac{1}{T_cC}\sum_{t,c}\widetilde{\Delta }_{t,c}^2 
\end{equation}

\begin{equation}
\widetilde{\Delta } _{t,c}^2=\left\{\begin{matrix}\Delta  _{t,c}  \quad , \Delta _{t,c}\le \tau \\\quad \tau \quad ,otherwise \end{matrix}\right.
\end{equation}

\begin{equation}
    \Delta_{t,c}=\left | log\widetilde{y}_{t,c} - log\widetilde{y}_{t-1,c}  \right |
\end{equation}

$C$ represents the activity category, $T_c$ denotes the number of samples belonging to the c-th category, and $\widetilde{y}_{t,c}$ signifies the probability of the activity category c being represented by the patch at time t and $\tau$ is the smoothing threshold($\tau = 2$).

For the activity label sequence $\hat{y}$ output by the model, in order to reduce the impact of over-segmentation on subsequent activity merging, we set each activity label as the center, set a smoothing window with a size of smooth\_size, and set the starting position to the Half of the activity label sequence within the window. Within the window, we count and sort the activities of each category, and select the activity category with the most surrounding distribution as the current activity label according to the sorting result. After updating the activity label sequence, the performance of activity recognition and segmentation. Finally, it is enhanced by linking Patches of the same activity class to determine the start and end position of each activity in the data flow. We set the boundary set of each activity $B = \{(s_i, e_i)\}_{i= 1}^N$, where $s_i$ represents the start subscript of the \textit{i-th} activity, $e_i$ represents the end subscript of the \textit{i-th} activity, and \textit{N} is the number of activities in the activity label sequence. The activity set corresponding to each boundary is represented by $\hat{a} = \{a_i\}_{i=1}^N$, where $a_i$ is the activity corresponding to the \textit{i-th} boundary. The Process of Algorithm is described in the Algorithm 1.

\subsection{Activity forecasting}
Within the activity forecast module, the continuous input Patch blocks retain one Patch block at the end. This involves filling \textit{P} repeated values of the last $\mathbf{x}_i\in \mathbb{R}$ value at the end of the original sequence. The resulting sequence of generated Patches, $\mathbf{x}_{p}^{(i)}$, is then represented as $\mathbf{z}^{'(i)}\in \mathbb{R}^{D\times N}$ through the Transformer backbone. By utilizing a linear layer and a fully connected layer with Softmax as the activation function, we obtain the predicted activity sequence, $\hat{y}_{pre}={\hat{y}_{L+1},\cdots,\hat{y}_{L+T_{P}}}\in \mathbb{R}^{1 \times T_P}$, using the representation of the last patch. It is important to note that $T_P$, the length of the future segment, is determined by the number of Patch blocks within the time $T_{pre}$, specifically $T_p=\frac{T_{pre}}{P}$.

\begin{algorithm}[htb]
	\caption{Segmentation Algorithm}\label{Prediction}
	\renewcommand{\algorithmicrequire}{\textbf{Input:}}
	\renewcommand{\algorithmicensure}{\textbf{Output:}}
	\begin{algorithmic}[1]
		\REQUIRE   
		Sensor dataset: $\bf{X}$ \\
		\ENSURE
            Activity label sequence segmentation result: $\hat{y}$\\
            \hspace{0.75cm}Activity boundaries: $B = \{(s_i, e_i)\}_{i=1}^N$\\
            \hspace{0.75cm}Activity class: $\hat{a} = \{a_i\}_{i=1}^N$\\ 
		\STATE $\hat{y} \leftarrow Model(\bf{X})$ by Eq.(1), (2), (3), (4) and Eq(5) \\	
            \STATE $smooth\_size$ $\leftarrow$ smoothing window size\\
            \STATE set a smooth sequence of active label: smooth$\_$data $\leftarrow$ []
            \STATE $half\_size \leftarrow smooth\_size // 2$
            \FOR{$i = 0$ to \textit{$len(\hat{y})$} }
            \STATE start $\leftarrow$ max(0, $i - half\_size$)\\
            \STATE end $\leftarrow$ min(len($\hat{y}$, $i + half\_size + 1)$)
            \STATE new\_activity\_label $\leftarrow$ cal$\_$max(sort($\hat{y}[start:end]$))
            \STATE add new\_activity\_label to smooth$\_$data
            \ENDFOR
            \STATE $\hat{y}$ $\leftarrow$ smooth$\_$data
		\STATE current activity type a $\leftarrow $ $\hat{y}[0]$ \\
            \STATE initial activity set index $j$
            \STATE activity start and end index s, e $\leftarrow$ 0
            \FOR{$i$ $=$ $1$ to \textit{$len(\hat{y})$} }
            \IF{$\hat{y}_{i}$ $\neq$ a}
		\STATE Add $a$ to $\hat{a}_{j}$
            \STATE $a$ $\leftarrow$ $\hat{y}_{i}$
            \STATE $e$ $\leftarrow$ $i$
		\STATE Add ($s$, $e$) to $B$\\
		\STATE $s$ $\leftarrow$ $e$ + 1 
            \STATE $j$ $\leftarrow$ $j$ + 1 
		\ENDIF
\ENDFOR
	\end{algorithmic}
\end{algorithm}
In the training stage, based on the predicted activity label sequence $\hat{y}_{pre}$ and the ground-truth activity sequence $y_{gt}$, all patch-level prediction losses can be expressed as:

\begin{equation}
    L_{pre}=l_{p}(\hat{y}_{pre},y_{gt})
\end{equation}

Finally, the previously mentioned loss functions from Equations (6), (8), and (11) are incorporated into the equation, resulting in the final loss function:

\begin{equation}
    L=L_{cls} + L _{seg} + L_{pre}
\end{equation}
We have introduced the main structure of P2LHAP. Algorithm 2 summarizes the training process of P2LHAP.

\begin{algorithm}[htb]
	\caption{The training process of P2LHAP}\label{Prediction}
	\renewcommand{\algorithmicrequire}{\textbf{Input:}}
	\renewcommand{\algorithmicensure}{\textbf{Output:}}
	\begin{algorithmic}[1]
		\REQUIRE   
		Training dataset: $\{\textbf{X}_{train}, y_{train}\}$ \\
            \hspace{0.4cm} Validation dataset: $\{\textbf{X}_{val}, y_{val}\}$\\
            \hspace{0.4cm} Test dataset: $\{\textbf{X}_{test}, y_{test}\}$\\
		\ENSURE
            Patch-level activity label predictions: $\hat{y}$\\
            \STATE Initialize training epoch e,
$\text{optimizer} \leftarrow \text{Adam}$
            \WHILE{$ e \le  Epoch_{max}$ and $ Convergent $}
            \STATE Get a mini-batch data $\textbf{X}_{batch}$ in $\textbf{X}_{train}$
            \STATE $\hat{y} \leftarrow Model(\textbf{X}_{batch})$ by Eq.(1), (2), (3), (4) and Eq.(5) \\
            \STATE $y_{gt} \leftarrow y_{train}$\\
            \STATE $\hat{y}_{pre}$ $\leftarrow$ $\hat{y}[-T_{p}:-1]$\\
            \STATE Calculate $L_{cls}$ with eq.(6), (7)\\
            \STATE Calculate $L_{pre}$ with eq.(11)\\ 
            \STATE Get $\hat{y}_{seg}$ through algorithm 1\\
            \STATE $\hat{y}$ $\leftarrow$ $\hat{y}_{seg}$\\
            \STATE Calculate $L_{\text{seg}}(\hat{y}$, $y_{\text{gt}}$) with eq.(8)\\
            \STATE e $\leftarrow$ e + 1\\
            \STATE Update model parameters based on eq.(12)\\
            \STATE $adjust\_learning\_rate(optimizer, e, args)$
            \STATE Calculate $\hat{y}_{val}$ with Eq.(6), (8)  and check convergence\\
            \ENDWHILE
		\STATE Update $\hat{y}$ from test sequence $\textbf{X}_{test}$ based on the model	
            
	\end{algorithmic}
\end{algorithm}

\section{EXPERIMENTS AND RESULTS}
In this section, we provide a comprehensive overview of our experiments and utilize cutting-edge research to evaluate our proposed multi-task framework. Firstly, we present the experimental settings, datasets, and evaluation metrics employed in our study. Subsequently, we perform a series of experiments to assess the effectiveness of our method, including qualitative and quantitative overall performance analysis as well as ablation experiments conducted using the Patch strategy.

\subsection{Experimental setup}
Our method was implemented using PyTorch and trained on an RTX 3090 GPU. The batch size was set to 64, and we utilized the Adam optimizer with a learning rate of 0.0001. The fully connected dropout rate is set at 20\% (0.2), head dropout is fixed at 5\% (0.05), the default patch length is 10, with a step size of 10. The model employs 8 heads by default, consists of 6 encoder-decoder layers, and undergoes training for 100 epochs.

For classification, segmentation, and forecast tasks, we employed the entire continuous dataset as input and divided it into training, validation, and test sets with a ratio of 0.7, 0.1, and 0.2 respectively. To evaluate the classification and segmentation performance of the model, we compared the forecast output with the real label. In the forecast task, the forecast patch length was set to 8, and performance assessment was based on a comparison between the forecast output patch label and the real label.

The choice of several parameters values for the models is shown in the table \ref{parm_table}:

\begin{table}[htbp]
	\setlength{\tabcolsep}{7mm}
        \renewcommand\arraystretch{1.5}
	\centering 
	\caption{\label{parm_table}The default choice of several parameters values for the models}
	\resizebox{\linewidth}{!}{
	\begin{tabular}{lcc}
		\toprule
		Model Parameters & Parameters Values & Parameter Introduction \\
		\midrule
		seq\_len & 200/151/512 & \parbox{4cm}{\centering input sensor sequence length 200(WISDM)/151(Unimib SHAR)/512(PAMAP2)}\\
		pred\_len & 8 & prediction label sequence length \\
        fc\_dropout & 0.2 & fully connected dropout \\
        attn\_dropout & 0.05 & attention dropout\\
        head\_dropout & 0.00 & transformer head dropout\\
        patch\_len & 10 & patch length\\
        stride & 10 & stride\\
        revin & 1 & RevIN True: 1 False: 0\\
        enc\_in & 3/3/9 & \parbox{4cm}{\centering encoder input size(the number of sensor channels) 3(WISDM)/3(Unimib SHAR)/9(PAMAP2)}\\
        dec\_in & 3/3/9 & \parbox{4cm}{\centering decoder input size(the number of sensor channels) 3(WISDM)/3(Unimib SHAR)/9(PAMAP2)}\\
        n\_heads & 8 & number of transformer heads\\
        e\_layers & 6 & number of encoder layers\\
        d\_layers & 6 & number of decoder layers\\
        d\_ff & 256 & dimension of FCN\\
        class\_num & 6/17/12 & number of activities\\
        d\_model & 256 & dimension of transformer model\\
        s\_t & 2 & smoothing threshold\\
        d\_k & d\_model//n\_heads & dimension of the query/key vectors per attention head\\
        d\_v & d\_model//n\_heads & dimension of the value vectors per attention head\\
        
		\bottomrule
	\end{tabular}
    }
\end{table}

1) $\textit{Evaluation metrics: }$ For the evaluation of the obtained Patch-level label sequences, we use the following metrics:

\textbf{1. Activity classification}

\textbf{Accuracy:} Accuracy is the overall measure of correctness across all classes and is calculated as follows:
\begin{equation}
    Accuracy=\frac{\sum_{c=1}^{C}TP_{c}+\sum_{c=1}^{C}TN_{c}}{\sum_{c=1}^{C}(TP_{c}+TN_{c}+FP_{c}+FN_{c})}
\end{equation}

\textbf{Weighted F1 score:} The weighted F1 score is calculated based on the proportion of Patch samples:

\begin{equation}
    F_w=\sum_{c=1}^{C}2*w_c\frac{prec_c\cdot recall_c}{prec_c+recall_c}
\end{equation}

Where $prec_c=\frac{TP_c}{TP_c+FP_c}$ and $recall_c=\frac{TP_c}{TP_c+FN_c}$ represent precision and recall, respectively. The weight $w_c$ is calculated as the ratio of $n_c$ to $N$, where $n_c$ is the number of Patch samples per class, and $N$ is the total number of Patch samples.

\textbf{2. Activity segmentation}

\textbf{Jaccard Index:} The proposed metric assesses the degree of overlap between actual activity segments and predicted activity segments, specifically focusing on the sensitivity towards over-segmentation errors. For the purpose of representation, $G_i$ and $P_i$ are introduced as respective variables denoting the actual and predicted activity ranges. In this context, the subscript i signifies the ith segment within class C. The calculation of this metric can be determined utilizing the prescribed formula:

\begin{equation}
    JI=\frac{1}{C}\sum_{c=1}^{C}\frac{G_i\cap P_i}{G_i\cup P_i}   
\end{equation}



\textbf{3. Activity forecast}

\textbf{Mean Square Error(MSE):} MSE serves as a prevalent metric to evaluate the performance of forecast models. It computes the average of the squared differences between predicted values and actual values, thereby providing a quantification of the degree of deviation between the predicted and actual values. A lower MSE signifies a more precise alignment between the model's predictions and the actual values, indicating superior performance. The formula for calculating MSE can be represented as follows:

\begin{equation}
    MSE = \frac{1}{n}\sum_{i=1}^{n}(y_{i}-\hat{y}_{i})^{2}
\end{equation}

\subsection{Datasets}

To evaluate our proposed method, we adopt three challenging public HAR datasets: WISDM, PAMAP2, UNIMIB SHAR. A brief introduction to the dataset follows.

\textbf{WISDM Dataset\cite{kwapisz2011activity}:} The data were obtained by 29 participants using phones with triaxial acceleration sensors placed in their trouser pockets with a sampling frequency of 20 Hz. Walking, strolling, walking up stairs, walking down stairs, standing motionless, and standing up were among the six daily activities undertaken by each participant. Fill missing values in a dataset using linear interpolation.

\textbf{PAMAP2 Dataset\cite{reiss2012introducing}:} This dataset comprises data collected from nine participants wearing IMUs on their chest, hands, and ankles with a sampling frequency of 100 Hz. These IMUs gather information on acceleration, angular velocity, and magnetic sensors. Each participant completed 12 mandatory activities, including lying down, standing, and going up and down stairs, as well as 6 optional activities, such as watching TV, driving a car, and playing ball.

\textbf{UNIMIB SHAR Dataset\cite{micucci2017unimib}:}  The dataset was collected from the University of Milano-Bicocca. Thirty volunteers, aged 18 to 60, had their Samsung phones equipped with Bosch BMA 220 sensors with a sampling frequency of 50 Hz. The study recorded 11,771 activities by placing phones in the participants' left and right pockets. The dataset comprises 17 categories of activities, divided into two groups: 9 activities of daily living (ADL) and 8 falling behaviors. Each activity was repeated 3 to 6 times.

\subsection{Comparisons with activity recognition algorithm}

In this section, we compare our method with other competing methods that have been evaluated on three publicly available datasets for wearable-based Human Activity Recognition. To conduct a thorough performance analysis, we use non-overlapping patches of size 10 to compare the overall recognition performance.

We compared the recognition performance of our proposed method with other similar algorithms as mentioned earlier. The results of various competing methods are summarized in Table \ref{activity recognition result}. Our method demonstrated improvements in recognition accuracy across three publicly available HAR datasets. Specifically, on the Unimib and PAMAP2 datasets, our method achieved the highest recognition performance, obtaining weighted average F1 scores of 82.04\% and 98.92\% respectively. Remarkably, our method surpassed the performance of the leading competing method by 3.26\% in terms of weighted average F1 score on the Unimib dataset.
        
        
\begin{table*}[htbp]
    \renewcommand\arraystretch{1.5}
    \centering
    \caption{\label{activity recognition result} The comparison of the accuracy and F1 performance of our work with other works on three datasets.}
    \begin{tabular}{ccccccccc}
        \toprule
        
          \multicolumn{3}{c}{PAMAP2} & \multicolumn{3}{c}{WISDM} & \multicolumn{3}{c}{UNIMIB SHAR} \\
         \cmidrule(r){1-3} \cmidrule(r){4-6} \cmidrule(l){7-9} 
         methods & Acc & F1 & methods & Acc & F1 & methods & Acc & F1 \\
        \midrule
        ours &\textbf{98.71\%} & \textbf{98.92\%} & ours &\underline{97.51\%} & 97.52\%   & ours & \textbf{81.51\%} & \textbf{82.04\%} \\
        
        \hline
        Y Li et al. (2023)\cite{li2023temporal} &\underline{98.04\%} & -  &K. Xia et al. (2020)\cite{xia2020lstm}& 92.71\% & 92.63\% &Y. Tang et al. (2022) \cite{tang2022triple} &76.30\% & 75.38\% 
		\\
           W. Gao et al. (2021)\cite{gao2021deep}&93.80\%  & 93.38\% & S Kobayashi et al. (2023)\cite{kobayashi2023marnasnets} & 90.62\% & - & W. Gao et al. (2021) \cite{gao2021deep} & 75.89\% & 74.63\% 
		\\
       SK. Challa et al. (2022)\cite{challa2022multibranch} &94.27\%  & 94.25\% &Z. N. Khan et al. (2021)\cite{khan2021attention}& 96.85\% & 97.20\% &F. Duan et al. (2023)\cite{duan2023multi}& 76.48\% & 75.71\% 
        		\\
           Al-Qaness et al. (2022)\cite{al2022multi}&93.19\%  & 92.96\% &W. Gao et al. (2021)\cite{gao2021deep}& 97.51\% & 97.25\%  &Al-Qaness et al. (2022)\cite{al2022multi} & 77.29\% & 76.83\% 
          \\
         F. Duan et al. (2023)\cite{duan2023multi} &94.50\%  & 94.80\% &W. Huang et al. (2023)\cite{9774930}& \bf{99.04\% }& \bf{99.18\%}  & W. Huang et al. (2023)\cite{9774930} &78.65\% & 78.78\% 
          \\
          W. Huang et al. (2023) \cite{9774930}&92.14\%  & 92.18\% &J. Li et al. (2023)\cite{10040612}& 97.11\% & \underline{98.64\%}  &Y. Tang et al. (2022) \cite{tang2022multiscale}  & \underline{79.02\%} & \underline{79.19\% }   \\
          S. Xia et al. (2022)\cite{xia2022boundary} &94.59\%  & 94.47\%  &SK. Challa et al. (2022)\cite{challa2022multibranch}&96.04\% & 96.05\%   & && 
          \\
           J. Li et al. (2023) \cite{10040612}&97.39\%  & 98.12\% & MK Yi et al.(2023)\cite{yi2023human}& 96.07\% &96.14\% && & 
          \\
          Y. Wang et al. (2023)\cite{10099032}& -  & \underline{98.33\%} &&  & && & 
          \\
    \midrule
        \bottomrule
    \end{tabular}
\end{table*}
\begin{figure}[htpb]
	\centering
	\subfloat[]{\includegraphics[width=0.45\textwidth]{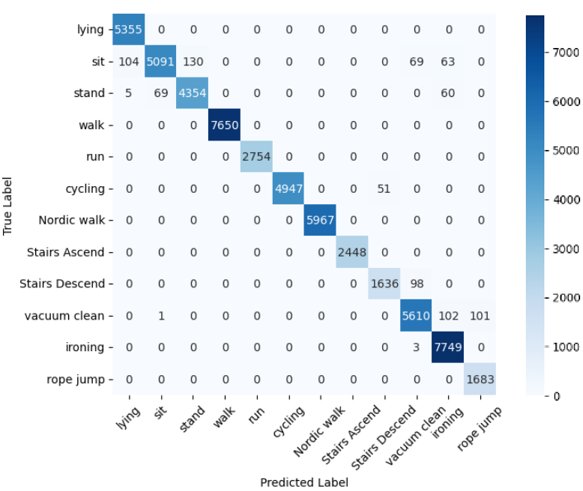}}
	
	\subfloat[]{\includegraphics[width=0.45\textwidth]{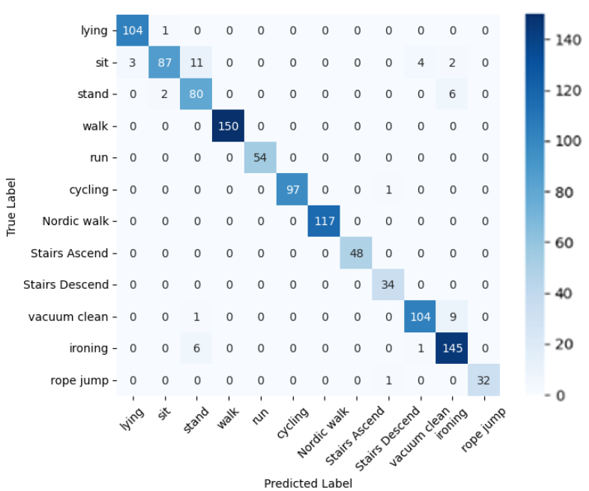}
		}
	\caption{The confusion matrices on the PAMAP2 dataset for patch sizes of 10 and 200 (a) $Patch\_size$=10, (b) $Patch\_size$=200}
	\label{confusion_pamap2}
\end{figure}
In Fig. \ref{confusion_pamap2}, we present the confusion matrix of our method for activity classification on the PAMAP2 dataset using patch sizes of 200 and 10, respectively. The confusion matrix provides insight into the accuracy of our method for various activity categories. It is evident that, despite the different patch sizes, there is minimal variation in overall accuracy. However, when the patch size is small, over-segmentation occurs, necessitating active segmentation through dividing the patch blocks. This approach enhances the recognition performance for most activity categories.
\begin{figure*}[htpb]
	\centering
	\includegraphics[width=1.0\textwidth]{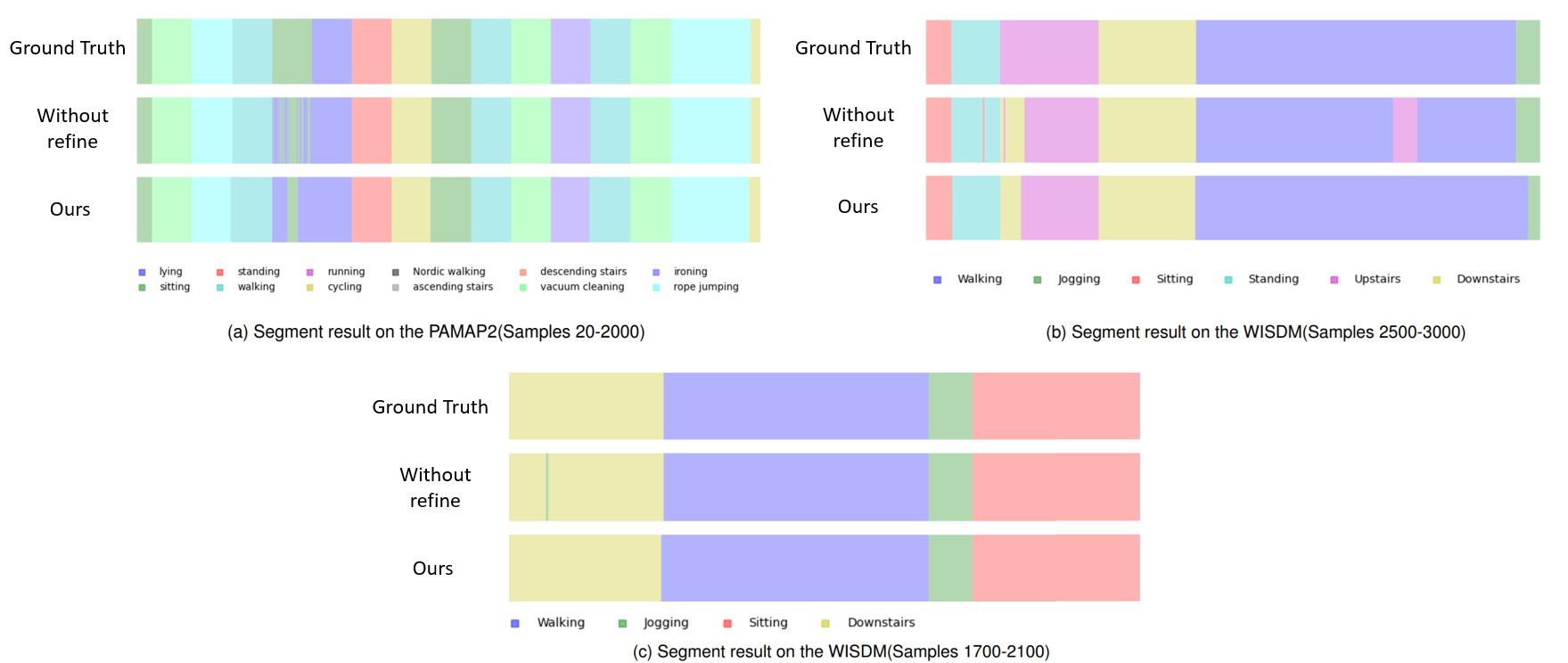}%
	\caption{Segment visualization results on PAMAP2 and WISDM datasets}
	\label{segment visualization comparsion}
\end{figure*}

\subsection{Qualitative segmentation results}

To demonstrate our proposed method's segmentation performance, this section presents a visualization of patch-level activity predictions in Fig. \ref{segment visualization comparsion}, with each color representing a different activity category. We compare sequence fragments obtained through model segmentation with selected sequence fragments from the PAMAP2 and WISDM datasets. In practical application scenarios, excessive segmentation can adversely affect prediction accuracy. Nonetheless, as depicted in the figure, our method successfully reduces most over-segmentation errors, leading to more precise and consistent patch-level predicted segments within each active segment.

Table \ref{dynamic segmentation result} displays the segmentation performance of our proposed method in comparison with competing algorithms. Our method excels in both segmentation and recognition tasks, consistently outperforming the comparison algorithms. Notably, our Jaccard Index scores on the Pamap2 dataset surpassed those of A\&D and JSR by 9.82\% and 3.47\% respectively, where (w/o) represents the segmentation Jaccard Index scores obtained without using the smoothing strategy. This justifies our employment of a smoothing algorithm, which effectively mitigates the boundary irregularities resulting from excessive segmentation. The algorithm serves to enhance clarity by refining the boundaries and ensures clearer and more distinct segmentation for each individual activity.

Sensor-recorded human activity data forms a continuous data stream, encompassing diverse actions like standing, sitting, walking, and transitions such as stand-to-sit and sit-to-lying. These activities are classified into two categories: basic and transition, based on their duration. Basic activities, characterized by longer periods, can be either dynamic (e.g., walking) or static (e.g., sitting), whereas transitional activities involve brief actions (seconds), like posture shifts (e.g., sit-to-stand). Despite the prevalence of research on basic activities, transition events are frequently overlooked due to their lower frequency and shorter duration compared to basic activities. When considering the short-lived activities that intervene between consecutive basic activities, two distinct scenarios may arise:
\begin{itemize} 
    \item When two consecutive basic activities differ, the activity segments in between are classified as transition activities. 
    \item If adjacent basic activities are identical, the segments in between are designated as the disturbance process of the activity. This occurs due to external influences or human factors disrupting the sensor data, causing irregular fluctuations over time. Consequently, the activities preceding and following the disturbance are considered of the same type.
\end{itemize}

Given the brief nature of transition activities, our framework efficiently divides the original sensor data into multiple patches, ensuring that the information of these short events can be effectively captured and learned. As demonstrated in Fig. 4(c), the short duration of the Jogging activity does not hinder our framework's ability to accurately segment it, showcasing its effectiveness.

\begin{table}[htbp]
	\renewcommand\arraystretch{1.5}
	\centering 
	\caption{\label{dynamic segmentation result} RESULTS OF SEGMENTATION METRICS ON PAMAP2 DATASET}
	\begin{tabular}{lccccc}
		\toprule
		\multirow{2}{*}{\bf{Seg Metric}} & \multicolumn{5}{c}{PAMAP2 DATASET} 
  \\ \cmidrule(r){2-6}
		 & Attn  & A\&D & JSR & ours(w/o) & Ours\\
   
		\midrule
    Jaccard Index&0.7908 &0.8312 &0.8947 & 0.8445 &\bf{0.9294}\\
		\bottomrule
	\end{tabular}
\end{table}

\subsection{Patch forecast analysis}

\begin{figure*}[htpb]
	\centering
	\includegraphics[width=1.0\textwidth]{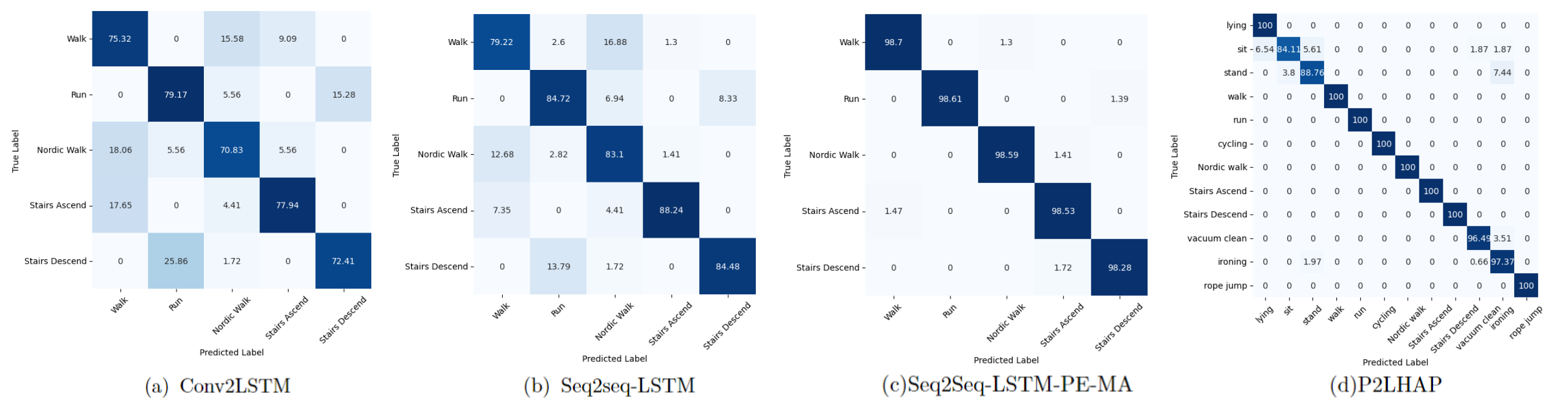}%
	\caption{ Confusion matrix of our model and three other models in predicting activity classification results on the PAMAP2 dataset}
	\label{predict visualization comparsion}
\end{figure*}

In this section, we assess the performance of the activity forecast module using our model to forecast the duration of activity categories in the next 8 patch blocks. Fig. \ref{predict visualization comparsion} displays the confusion matrix for the future activity forecast in the pamap2 dataset. Fig. \ref{predict visualization comparsion} (a),(b),(c) present the prediction confusion matrices of Conv2LSTM\cite{CANIZO2019246}, Seq2Seq-LSTM\cite{WangSeq2SeqLSTM}, and Seq2Seq-LSTM-PE-MA\cite{jaramillo2023human} predictors for five activity categories in the dataset, while Fig. (d) illustrates the predictions of our proposed model. The confusion matrix results for all campaigns indicate an average accuracy of 96.59\%. Additionally, we compare the predicted outputs of future sensor signals with the ground truth signals by evaluating their mean squared error (MSE). To enable the model to predict future sensor signals, we exclude the last classification header and utilize the first half of the input sensor sequence for model input, reserving the second half to match the size of the predicted output data. Table \ref{predict evaluation result} presents the performance comparison of our model with five forecasters for sensor sequence signal prediction.

\begin{table}[htbp]
    \setlength{\tabcolsep}{1mm}
	\renewcommand\arraystretch{1.5}
	\centering 
	\caption{\label{predict evaluation result}Evaluation results of the forecast on PAMAP2 dataset}
	\begin{tabular}{lccccc}
		\toprule
		\multirow{2}{*}{\bf{Metric}} & \multicolumn{5}{c}{Models} 
  \\ \cmidrule(r){2-6}
		 & SynSigGAN & Conv2LSTM & Seq2Seq-LSTM& TTS-GAN & Ours\\
   
		\midrule
    MSE & 0.355&0.302&0.240 &0.163&\bf{0.126}\\
		\bottomrule
	\end{tabular}
\end{table}

\subsection{Ablation experiments}
In this section, we conduct two ablation experiments to investigate the influence of varying patch block sizes on model performance and the efficacy of patching operations in generating label sequences.

\textbf{Impact of Patch Size:} various sizes and quantities of patches can yield different outcomes. Hence, we employ a scaling approach with multiple patches to analyze their performance. We provide the comprehensive results in Table \ref{patch size impact}. We observed a decrease in accuracy when the Patch size is reduced significantly. This decline can be attributed to the issue of excessive segmentation. However, when the Patch size is set to 10, it consistently exhibits exceptional recognition performance across all three datasets.

\textbf{The impact of Patching modules:} we examined the effects on the results by not patching the sensor data stream, but instead using the Transformer backbone directly. The F1 score obtained on the benchmark dataset is presented in Table \ref{no patching module}. It was observed that when the data was not patched during processing, the accuracy of the experimental results deteriorated. Hence, applying data patching enhanced the performance of activity recognition.

\textbf{The impact of the number of multiple heads and the number of encoder-decoder layers:}this study investigates the influence of model components on classification outcomes by varying the number of multi-heads and encoder-decoder layers. Table \ref{mheads impact} presents the activity recognition accuracy of our proposed framework on the PAMAP2 dataset, showcasing the performance changes with different configurations. The results indicate that an increase in both multi-heads count and model layer depth facilitates better learning of activity sequence representations, leading to enhanced recognition accuracy. The optimal configuration is achieved when the model has 6 layers and 8 multi-heads, achieving the highest accuracy.

\begin{table}[htbp]
	\renewcommand\arraystretch{1.5}
	\centering 
	\caption{\label{patch size impact}  $F_{1}$ values with different patch sizes on benchmark datasets}
	\begin{tabular}{p{2.6cm}ccccc}
		\toprule
		\diagbox{\bf{Datasets}}{\bf{Patch Size}}& P=1 & P=2 & P=5 &P=10 & P=20\\
		\midrule
        \quad \quad \quad WISDM&0.8544&0.9056&0.8819 &\bf{0.9751}&0.9568\\
		\quad \quad \quad PAMAP2 &0.9151&0.9406& 0.9610 &\bf{0.9892} & 0.9667\\
		\quad \quad \quad UNIMIB & 0.6719 & 0.6950 & 0.7566 & \bf{0.8204} & 0.7876 \\
		\bottomrule
	\end{tabular}
\end{table}

\begin{table}[htbp]
	\setlength{\tabcolsep}{4mm}
	\centering 
	\caption{\label{no patching module}Activity classification F$_{1}$ value without Patching module on benchmark datasets}
	\begin{tabular}{lccc}
		\toprule
		Model & WISDM & PAMAP2 & UNIMIB\\
		\midrule
		only transformer backbone & 0.9090 & 0.8783 & 0.7533 \\
            \midrule
            P2LHAP & \bf{0.9752} & \bf{0.9892} & \bf{0.8204} \\
		\bottomrule
	\end{tabular}
\end{table}

\begin{table}[htbp]
	\renewcommand\arraystretch{1.5}
	\centering 
	\caption{\label{mheads impact}  Activity recognition accuracy with different numbers of multi-heads and different numbers of encoder-decoder layers on the PAMAP2 dataset}
	\begin{tabular}{lccccc}
		\toprule
		\diagbox[width=10em]{\bf{Layer}}{\bf{Mutil-heads}}& 1 & 2 & 4 & 8 & 16\\
		\midrule
        \quad \quad \quad 1&0.8964&0.9259&0.9372 &0.9447&0.9485\\
		\quad \quad \quad 2 &0.9406&0.9556&0.9560 &0.9654 & 0.9581\\
		\quad \quad \quad 3 & 0.9693 & 0.9702 & 0.9633 & 0.9750 & 0.9719 \\
  \quad \quad \quad 6 & 0.9658 & 0.9741 & 0.9793 & \bf{0.9871} & 0.9788 \\
  \quad \quad \quad 8 & 0.9710 & 0.9793 &  0.9790 & 0.9802& 0.9812 \\
		\bottomrule
	\end{tabular}
\end{table}

\subsection{Further Analysis}

To thoroughly assess the practicality of our proposed methodology, we will examine its computational complexity and parameter count, aiming to determine its feasibility for real-time deployment on mobile or embedded devices. The computational analysis was conducted on a Linux platform equipped with an RTX 3090 GPU and an Intel Core i9-13900K. The experiments employed a batch size of 32 for both training and testing, with 3 encoder and decoder layers, and 4 multi-heads. Utilizing the PAMAP2 dataset, we set the input sequence length to 512, patch size to 24, and step size to 12.

The computational complexity comparison, as presented in the table \ref{complexity analysis}, reveals that our method exhibits superior parameter efficiency and lower inference time compared to the benchmarked techniques.

\begin{table}[htbp]
	\setlength{\tabcolsep}{4mm}
	\centering 
	\caption{\label{complexity analysis}Complexity comparison of several competing methods on the PAMAP2 dataset}
	\begin{tabular}{lccc}
		\toprule
		Methods & Param & FLOPs & Inference time(s/epoch)\\
		\midrule
		Attn.\cite{murahari2018attention}& 1.983M & 20.058M & 8.22 \\
            A\&D \cite{abedin2021attend} & 1.503M & 25.612M & 11.88 \\
            JSR \cite{xia2022boundary} & 2.799M & \bf{12.105M} & 0.78 \\
            Ours & \bf{1.449M} & 20.014M & \bf{0.08} \\
		\bottomrule
	\end{tabular}
\end{table}


\section{Discussion and Future Work}
This section delves into the merits and drawbacks of our proposed framework, with a focus on addressing the challenges posed by the inherent uncertainty in natural human activity durations and the complexities of IMU signal analysis for activity segmentation, recognition, and forecast. To tackle these issues, we introduce a novel multi-task deep learning framework that employs the transformer's Patch Sequence to Label Sequence architecture. The transformer network captures inter-patch relationships, enabling it to predict future activity sequences based on past representations. However, our framework's reliance on a closed space assumption in traditional machine learning restricts its ability to handle unknown activities, as the softmax classification layer misclassifies them as known categories. Future work will concentrate on enhancing the architecture to accommodate activity recognition in novel environments, by addressing the issue of unknown activity recognition. Below we will discuss the points planned for improvement in subsequent research:

1) Given that our model employs a Softmax function for classification, its output space is confined, precluding the ability to reject unknown classes, as the predicted probabilities are normalized among all known training classes. Consequently, the model struggles to accurately identify novel activities. To address this limitation, we intend to adopt Deep Open Classification (DOC) \cite{shu2017doc} in our future work. Unlike softmax, DOC employs a "1-vs-Rest" layer with N Sigmoid functions for N known classes, which serves as a basis for representing both known and unknown classes. By transferring the learned similarity knowledge from known classes to potential unknowns, we can leverage hierarchical clustering algorithms to cluster rejected instances based on the transferred similarity, thus revealing potential activity classes within the rejected samples.

2) Our model currently adopts a supervised training approach. Future iterations will incorporate self-supervision for retraining, as it eliminates the need for costly labeled datasets in real-world activity recognition scenarios. Self-supervised learning generates training targets autonomously and exploits unlabeled data to learn representations, addressing the challenge of limited labeled data or high annotation costs. This adaptation enhances the model's applicability to practical tasks. Moreover, by facilitating the learning of more universal and general feature representations for unseen activities, self-supervision contributes to improving the model's generalization across diverse activity recognition tasks.
\section{CONCLUSION}

In this paper, we propose P2LHAP, an activity-perception model that offers a patch sequence to label sequence model design solution tailored to identify, segment, and forecast human activities through sensor data analysis. Unlike conventional Transformer-based models, P2LHAP partitions the native sensor data stream into multiple patch blocks and utilizes a channel-independent approach to mitigate noise interference among sensor channels. The forecast of forthcoming activity sequences is achieved through reasoning with a Transformer encoder-decoder. To address the issue of over-segmentation, we refine labels based on the nearby distribution of labels close to the active label, thus reducing clutter in the distribution of active labels. Leveraging the P2LHAP framework, we effectively tackle activity segmentation, recognition, and forecast within the realm of activity perception tasks. Our experimental evaluation on three standard datasets demonstrates that our method surpasses existing approaches in classification, segmentation, and forecast tasks.


\bibliographystyle{IEEEtran}


\vspace{11pt}

\end{document}